# On Probability Distributions Over Possible Worlds

Fahiem Bacchus*
Department of Computer Science
University of Waterloo
Waterloo, Ontario, Canada
N2l–3G1
fbacchus@watdragon.waterloo.edu

## Abstract

In *Probabilistic Logic* Nilsson uses the device of a probability distribution over a set of possible worlds to assign probabilities to the sentences of a logical language. In his paper Nilsson concentrated on inference and associated computational issues. This paper, on the other hand, examines the probabilistic semantics in more detail, particularly for the case of first order languages, and attempts to explain some of the features and limitations of this form of probability logic. It is pointed out that the device of assigning probabilities to logical sentences has certain expressive limitations. In particular, statistical assertions are not easily expressed by such a device. This leads to certain difficulties with attempts to give probabilistic semantics to default reasoning using probabilities assigned to logical sentences.

## 1 Introduction

Nilsson [1] describes a method of assigning probabilities to the sentences of a logic through a probability distribution over a set of possible worlds. Each possible world in this set is a consistent assignment of truth values to the sentences of the logic, and the set consists of all unique possible worlds. A probability distribution is placed over this set. Probabilities are then assigned to the sentences by giving each sentence a probability equal to the probability of the subset of possible worlds in which it is true.

Although this approach is unproblematic when applied to propositional languages, certain difficulties arise when dealing with first order languages. By taking a different tack these difficulties can be overcome, and indeed, it has already been demonstrated that probabilities can be coherently assigned to the sentences of any first order language (Gaifman [2], Scott and Krauss [3]).

While the method of assigning probabilities to logical formulas is capable of representing probabilistic degrees of belief, it is incapable of effectively representing statistical assertions. It is argued that many types of defaults have a natural statistical interpretation, but cannot be represented by probabilities over logical formulas, because of this limitation. Some authors have attempted to represent defaults by (conditional) probabilities over logical formulas (Geffner and Pearl [4], Pearl [5]), and the difficulties this causes can be demonstrated.

It is pointed out that although probabilities over logical formulas fails to do the job, statistical assertions can be efficiently represented in other types of probability logics, logics which go beyond the simple device of assigning probabili-

*This research was supported by a Post-Doctoral fellowship funded by the U.S. Army Signals Warfare Laboratory, while the author was a researcher at the University of Rochester.

ties to first order sentences.

## 2 The Propositional Case

A natural semantic model for a propositional language is simply a subset of the set of atomic symbols (Chang and Keisler [6]). This subset is the set of atomic symbols which are assigned the truth value true (t). In the propositional case Nilsson's concept of possible worlds, i.e., a set of consistent truth value assignments, has a natural correspondence with the set of semantic models. Each possible world is completely determined by its truth value assignments to the atomic symbols of the language, and the truth value assignments to the atomic symbols can be viewed as being the characteristic function of a semantic model (with $t = 1, f = 0$).

For example, in a propositional language with two atomic symbols $\{A, B\}$ there are four possible worlds with corresponding semantic models ($\sigma$ is used to indicate the truth function).

1. $\{A^\sigma = t, B^\sigma = t\}$ or $\{A, B\}$.
2. $\{A^\sigma = t, B^\sigma = f\}$ or $\{A\}$.
3. $\{A^\sigma = f, B^\sigma = t\}$ or $\{B\}$.
4. $\{A^\sigma = f, B^\sigma = f\}$ or $\{\}$, i.e., the empty set.

Another way of looking at possible worlds, which will turn out to be more useful when we move to first order languages, is to consider the *atoms*[1] of the language. When the language has a finite number of atomic symbols each possible world can be represented as a single sentence: a sentence formed by conjoining each atomic symbol or its negation, such a sentence is called an atom. Corresponding to the four worlds above we have the four atoms $A \wedge B$, $A \wedge \neg B$, $\neg A \wedge B$, and $\neg A \wedge \neg B$.

---

[1]An atom in a Boolean algebra is a minimal non-zero element (Bell and Machover, [7]).

Given a probability distribution over the set of possible worlds it is possible to assign a probability to each sentence of the language. Each sentence is given a probability equal to the probability of the set of worlds in which it is true. So, for example, the sentence $A \vee B$ is true in worlds 1, 2, and 3. Hence, its probability will be equal to the probability of the set of worlds $\{1, 2, 3\}$.

Equivalently, a probability distribution can be placed directly over the sentences of the logic, more precisely over the Lindenbaum-Tarski algebra of the language. This algebra is generated by grouping the sentences into equivalence classes. Two sentences, $\alpha$ and $\beta$, are in the same equivalence class iff $\vdash_0 \alpha \leftrightarrow \beta$, where $\vdash_0$ indicates deducible from the propositional axioms.

This technique is not limited to languages with a finite number of atomic symbols. However, when the language is finite the atoms will be sentences of the language, and the probability distribution can be completely specified by the probabilities of the atoms (the e-classes of). Any sentence can be written as a disjunction of a unique set of atoms, and its probability will be the sum of the probabilities of these atoms. For example, if we specify the probabilities $\{A \wedge B = .5, A \wedge \neg B = .1, \neg A \wedge B = .2, \neg A \wedge \neg B = .2\}$, then the sentence $A \vee B$ will have probability 0.8 as it can be written as $(A \wedge B) \vee (A \wedge \neg B) \vee (\neg A \wedge B)$.

## 3 First Order Languages

When the move is made to first order languages certain problems arise. The first problem is that we lose the nice correspondence between possible worlds and semantic models. The normal semantic model for a first order language is considerably more complex than the model for a propositional language, and the truth values of the sentences in a first order language are determined both by the model and by an interpretation (i.e., the mapping from the symbols to the semantic entities). For a given truth value assignment to the sentences (possible world) there will be many different (in fact an infinite number) of model/interpretation pairs which will yield the same truth values. Hence, the semantic structure of the possible worlds is unclear.

Another difficulty, which Nilsson is aware of, is that Nilsson's techniques depend on being able to generate consistent truth value assignments for a set of sentences. These are used as 0/1 column vectors in his $V$ matrix. This technique is limited to languages in which the consistency of a finite set of sentences can be established. The consistency of a set of first order sentences is not decidable, except in special cases (see Ackermann [8] for an interesting survey).

These difficulties can be avoided if instead of probability distributions over possible worlds we consider probability distributions over the Lindenbaum-Tarski (L-T) algebra of the language. It has already been demonstrated by Gaifman [2] that a probability measure can be defined over the L-T algebra of sentences of a first order language. Every sentence in the language will have a probability equal to the probability of its equivalence class, and furthermore, the probabilities will satisfy the condition

If $\vdash \neg(\alpha \wedge \beta)$ then $p[\alpha \vee \beta] = p[\alpha] + p[\beta]$,

where $\vdash$ indicates deducible from the first order axioms. This means that the probability measure preserves the partial order of the algebra. In this partial order we have $\alpha < \beta$ iff $\alpha \wedge \beta = \alpha$; hence, $p[\alpha] \leq p[\alpha \wedge \beta] + p[\neg\alpha \wedge \beta] = p[\beta]$ (by the above condition). Under this partial order the conjunction and disjunction operators generate the greatest lower bound (infimum) and least upper bound (supremum).

To examine what happens to quantified sentences under such a probability measure it is sufficient to note that for L-T algebras we have that

$(\star)\ \ |\exists x \alpha| = \bigvee_{t \in T} |\alpha(x/t)|,$

where $|\bullet|$ indicates the equivalence class of the formula, and $T$ is the set of terms of the language. What this means is that each existentially quantified sentence is equal to the supremum of all its instantiations. This implies that the probability of any existentially quantified sentence must be greater than or equal to the probability of any of its instances. Similarly, the probability of any universally quantified sentence must be less than or equal to the probability of any of its instances.

This interpretation also makes sense in terms of Nilsson's possible worlds. In any single possible world the existential must be true if any of its instantiations are. Hence, the set of possible worlds in which the existential is true includes the set of possible worlds in which any instantiation is true, and thus the existential must have a probability greater than or equal to the probability of any of its instances.

## 4 The Representation of Statistical Knowledge

Probabilities attached to logical sentences can be interpreted as degrees of belief in those sentences. Instead of either asserting a sentence or its negation, as in ordinary logic, one can attach some intermediate degree to it, a degree of belief. So, for example, one could represent a degree of belief of greater than 0.9 in the assertion "Tweety can fly" by assigning the sentence $Fly(Tweety)$ a probability $> 0.9$. However, it is not easy to represent statistical information, for example, the assertion "More than 90% of all birds fly."[2]

First, propositional languages do not seem to possess sufficient power to represent these kinds of statements. This particular statistical state-ment is an assertion which indicates some relationship between the properties of being a bird and being able to fly, but it is not an assertion about any particular bird. This indicates that some sort of variable is required. Propositional languages do not have variables, and so are inadequate for this task even when they are generalized to take on probabilities instead of just 1/0 truth values.

---

[2]It is the case that first order logic is in some sense universally expressive. That is, set theory can be constructed in first order logic, and thus sufficient mathematics can be built up inside the language to represent statements of this form. This is not, however, an efficient representation, nor is there any direct reflection in the semantics of the statistical information. I am concerned here with efficient representations.

When we move to first order languages we do get access to variables, variables which can range over the set of individuals. A seemingly reasonable way to represent this statement is to consider the probabilistic generalization of the universal sentence $\forall x Bird(x) \rightarrow Fly(x)$. The universal in 1/0 first order logic says that all birds fly, so if we attach a probability of $> 0.9$ perhaps we will get what we need. Unfortunately, this does not work. If there is single bird who is thought to be unable to fly, this universal will be forced to have a probability close to zero. That is, the probability of this universal must be $1 - p[\exists x Bird(x) \wedge \neg Fly(x)]$. Hence, if one believes to degree greater that 0.1 that a non-flying bird exists, then the probability of the universal must be $< .9$.

Since universal quantification or its dual existential quantification are the only ones available in a first order language, it does not seem that moving to first order languages allows us to represent statistical assertions. There is, however, one more avenue available: conditional probabilities. We have probabilities attached to sentences hence with two sentences we can form conditional probabilities. It has been suggested (Cheeseman [9]) that meta-quantified statements of the following form can be used to capture statistical statements, in particular for the statement about birds:

$$(\forall x) p[Fly(x)|Bird(x)] > 0.9.$$

The reason that this is a meta-quantification is that the universal quantifier is quantifying over a formula "$p[Fly(x)|Bird(x)]$" which is not a formula of a first order language (unconditional probability assertions like "$p[fly(x)]$" are not first order formulas either). This statement is intended to assert that for every term, $t$, in the first order language the conditional probability of the sentence $Fly(x/t)$, with the variable $x$ substituted by the term $t$, given $Bird(x/t)$ is $> 0.9$.

However, this formulation also falls prey to any know exception. Say that there is some individual, denoted by the constant $c$, who is thought to be a bird, i.e., $p[Bird(c)]$ is high, and for some reason or the other is also believed to be unable to fly, i.e., $p[Fly(c)]$ is low, then clearly this statement cannot be true for the instance when $x$ is $c$; hence, the meta-level universal statement cannot be true: it is not true for the instance $c$. It is important to note that it does not matter what other things are known about the individual $c$. For example, $c$ could be known to be an ostrich, and thus there may be a good reason why $c$ is unable to fly. However, it will still be the case that the conditional probability of $Fly(c)$ given just $Bird(c)$ will not be $> 0.9$. The universal statement will fail for $c$. That is, this problem is not resolved by conditioning on more knowledge as claimed by Cheeseman [9].

There is no way that the statistical statement "More than 90% of all birds fly" can be represented by the assertion that the conditional probability is greater than 0.9 for *all* substitutions of $x$: this assertion will be false for certain substitutions. The problem here is that the statistical statement implies that $p[Fly(x)|Bird(x)] > 0.9$ for a *random* $x$, but a universally quantified $x$ is not the same as a random variable $x$; furthermore, the simple device of assigning probabilities to sentences of a logical language does not give you access to random variables. This point has also been raised by Schubert [10].

One can choose to interpret the variable '$x$' as being a random variable. However, simply choosing such an interpretation is not sufficient; it does not provide any formal meaning. That is, the semantics and behaviour of such a random $x$ must also be specified. It is possible to formalize such an interpretation. A logic can be constructed with random variables which have a formal semantics and a proof theory which specifies their behaviour. Such a logic has been constructed (Bacchus [11,12]). However, its structure is quite different from probability logics which attach probabilities to first order sentences.

## 5 The Representation of Defaults

There are many different defaults which have a natural statistical justification, the famous example of "Birds fly" being one of them. A natural reason for assuming by default that a particular bird can fly is simply the fact that, in a statistical sense, most birds do fly. This is not to say that all defaults have a statistical interpretation: there are many different notions of typicality which do not have a straightforward statistical interpretation, e.g., "Dogs give live birth" (Carlson [13], Nutter [14], also see Brachman [15] for a discussion of different notions of typicality).

Since probabilities attached to the sentences of a logic do not offer any easy way of representing statistical assertions, it is not surprising that attempts to use this formalism to give meaning to defaults leads to certain difficulties.

Recently Geffner and Pearl [4] have proposed giving semantics to defaults through meta-quantified conditional probability statements (also Pearl [5][3]). For example, the default "Birds fly" is given meaning through the meta-quantified statement $\forall x \, p[Fly(x)|Bird(x)] \approx 1$ In order to allow penguins to be non-flying birds they have the separate default rule: $\forall x \, p[\neg Fly(x)|Penguin(x)] \approx 1$. They also have universal statements like $\forall x \, Penguin(x) \rightarrow Bird(x)$. The probability of these universals is one; thus, as discussed above, every instantiation must also have probability one.

[3]Pearl uses a slightly different notion of probabilities within $\epsilon$ of one. The technical differences between this approach and that of Geffner and Pearl do not make any difference to the following discussion; the anomalies presented also appear in Pearl's system.

To examine the difficulties which arise from this approach consider the following example. Say that we have a logical language with the predicates $Bird$, $Fly$, and $Penguin$, some set of terms $\{t_i\}$, and a probability distribution over the sentences of the language which satisfies the default rules, i.e., for all terms $t_i$, $p[Fly(t_i)|Bird(t_i)] \approx 1$ and $p[\neg Fly(t_i)|Penguin(t_i)] \approx 1$, and in which the universal $\forall x \, Penguin(x) \rightarrow Bird(x)$, has probability one. Some simple facts which follow from the universal having probability one are that for all terms $t_i$, $p[Bird(t_i)] \geq p[Penguin(t_i)]$, and $p[Bird(t_i) \wedge Penguin(t_i)] = p[Penguin(t_i)]$).

Consider the derivation in figure 1.

The constraints imply that for any term $t_i$ that

$$p[Penguin(t_i)] \leq (\approx) p[\neg Penguin(t_i)];$$

hence $p[Penguin(t_i)]$ cannot be much greater than 0.5. Since $\approx 0.5$ is an upper bound on the probability of all instances of the formula $Penguin(x)$, it must also be the case that it is an upper bound on the probability of the sentence $\exists x \, Penguin(x)$, by equation $\star$.

That is, if we accept the defaults we are must reject any sort of high level of belief in the *existence* of penguins.

To be fair to Geffner and Pearl their system does provide a *calculus* for reasoning with defaults. However, this result indicates that semantically their particular probabilistic interpretation of defaults causes anomalies. It does not capture the statistical notion that birds usually fly.

## 6 Conclusions

It has been demonstrated that although probabilities can be assigned to the sentences of any

19

$$\begin{aligned}
1 &\approx p[Fly(t_i)|Bird(t_i)] \\
&= \frac{p[Fly(t_i) \wedge Bird(t_i) \wedge \neg Peng(t_i)]}{p[Bird(t_i)]} + \frac{p[Fly(t_i) \wedge Bird(t_i) \wedge Peng(t_i)]}{p[Bird(t_i)]} \\
&\leq \frac{p[Fly(t_i) \wedge Bird(t_i) \wedge \neg Peng(t_i)]}{p[Penguin(t_i)]} + \frac{p[Fly(t_i) \wedge Bird(t_i) \wedge Peng(t_i)]}{p[Penguin(t_i)]} \\
&\leq \frac{p[\neg Penguin(t_i)]}{p[Penguin(t_i)]} + \frac{p[Fly(t_i) \wedge Penguin(t_i)]}{p[Penguin(t_i)]} \\
&= \frac{p[\neg Penguin(t_i)]}{p[Penguin(t_i)]} + \approx 0
\end{aligned}$$

Figure 1: Conditional Probabilities close to one.

first order language, the resulting probability logics are not powerful enough to efficiently represent statistical assertions. It has also been demonstrated that attempts to give defaults a probabilistic semantics using these types of probability logics leads to certain semantic anomalies.

Statistical facts, it has been argued, give a natural justification to many default inferences. This implies that probabilities might still be useful for giving semantics to default rules and a justification to default inferences. For example, the default rule "Birds fly" could be represented as a statistical assertion that some large percentage of birds fly, and the default inference "Tweety flies" could be given the justification that Tweety probably does fly if to the best of our knowledge Tweety was a randomly selected bird.

Probability logics which accomplish this have already been developed (Bacchus [11,12], Kyburg [16]), but these logics go beyond the simple device of assigning probabilities to the sentences of a logical language. Bacchus uses a logic which has random variables as well as universally quantified variables, this logic is capable of expressing statistical information, and possesses a sound and complete proof theory capable of reasoning with statistical facts. Default inferences are handled by an inductive mechanism which forms defeasible conclusions, conclusions which can be defeated by new information. Kyburg uses an object language/meta-language formalism, and has explored the inductive formation of defeasible conclusions in greater detail [17].

## 7 Acknowledgement